\documentclass{ecai} 
\usepackage{latexsym}
\usepackage{amssymb}
\usepackage{amsmath}
\usepackage{amsthm}
\usepackage{booktabs}
\usepackage{enumitem}
\usepackage{graphicx}
\usepackage{color}
\usepackage{listings}
\usepackage{times}
\usepackage{soul}
\usepackage{url}
\usepackage[hidelinks]{hyperref}
\usepackage[utf8]{inputenc}
\usepackage{graphicx}
\usepackage{amsmath}
\usepackage{amsthm}
\usepackage{booktabs}
\usepackage[switch]{lineno}
\usepackage{graphicx}  
\usepackage{svg}  
\usepackage{multirow}
\usepackage{float}
\usepackage{CJK}
\usepackage{algorithm,algpseudocode}
\usepackage{graphicx}
\usepackage{amsmath}
\usepackage{soul}
\usepackage{xcolor}
\usepackage{soul}
\usepackage{url}
\usepackage[hidelinks]{hyperref}
\usepackage[utf8]{inputenc}
\usepackage[small]{caption}
\newcommand{\BibTeX}{B\kern-.05em{\sc i\kern-.025em b}\kern-.08em\TeX}

\begin{document}

%%%%%%%%%%%%%%%%%%%%%%%%%%%%%%%%%%%%%%%%%%%%%%%%%%%%%%%%%%%%%%%%%%%%%%%%

\begin{frontmatter}

%%% Use this command to specify your submission number.
%%% In doubleblind mode, it will be printed on the first page.

%%% Use this command to specify the title of your paper.

\title{Introspection of Thought Helps AI Agents}

%%% Use this combinations of commands to specify all authors of your 
%%% paper. Use \fnms{} and \snm{} to indicate everyone's first names 
%%% and surname. This will help the publisher with indexing the 
%%% proceedings. Please use a reasonable approximation in case your 
%%% name does not neatly split into "first names" and "surname".
%%% Specifying your ORCID digital identifier is optional. 
%%% Use the \thanks{} command to indicate one or more corresponding 
%%% authors and their email address(es). If so desired, you can specify
%%% author contributions using the \footnote{} command.
\author[A,B]{\fnms{Haoran}~\snm{Sun}\thanks{Email: 2022090916002@std.uestc.edu.cn.}\footnote{Equal contribution.}}
\author[A,C]{\fnms{Shaoning}~\snm{Zeng}\thanks{Corresponding Author. Email: zeng@csj.uestc.edu.cn.}}

\address[A]{Yangtze Delta Region Institute (Huzhou), University of Electronic and Science Technology of China, Huzhou, Zhejiang, China}
\address[B]{School of Information and Software Engineering, University of Electronic and Science Technology of China, Chengdu, Sichuan, China}
\address[C]{Zhejiang Chuangjiekedong Ltd., Huzhou, Zhejiang, China}

%%% Use this environment to include an abstract of your paper.

\begin{abstract}
AI Agents rely on Large Language Models (LLMs) and Multimodal-LLMs (MLLMs) to perform interpretation and inference in text and image tasks without post-training, where LLMs and MLLMs play the most critical role and determine the initial ability and limitations of AI Agents. Usually, AI Agents utilize sophisticated prompt engineering and external reasoning framework to obtain a promising interaction with LLMs, e.g., Chain-of-Thought, Iteration of Thought and Image-of-Thought. However, they are still constrained by the inherent limitations of LLM in understanding natural language, and the iterative reasoning process will generate a large amount of inference cost. To this end, we propose a novel AI Agent Reasoning Framework with Introspection of Thought (INoT) by designing a new LLM-Read code in prompt. It enables LLM to execute programmatic dialogue reasoning processes following the code in prompt. Therefore, self-denial and reflection occur within LLM instead of outside LLM, which can reduce token cost effectively. Through our experiments on six benchmarks for three different tasks, the effectiveness of INoT is verified, with an average improvement of 7.95\% in performance, exceeding the baselines. Furthermore, the token cost of INoT is lower on average than the best performing method at baseline by 58.3\%. In addition, we demonstrate the versatility of INoT in image interpretation and inference through verification experiments.

%We also present a new evaluation metric to indicate the construction effects, by which an average score of $92.95\%$ is obtained. We also present a new evaluation metric to indicate the construction effects, by which an average score of $92.95\%$ is obtained.
\end{abstract}

\end{frontmatter}

\section{Introduction}

With the advancement of Large Language Models (LLM) and Multimodal LLMs (MLLM), from GPT-3 to modern models such as DeepSeek-V2.5 \cite{deepseek}, GPT-4 \cite{chatgptreview}, LLaVA \cite{llava} and Claude \cite{claude}, multimedia reasoning tasks has undergone a significant transformation. LLMs are widely applied in various reasoning tasks \cite{codegenerate,imageofthought}, particularly in mathematical reasoning \cite{chatgptreview}, programming \cite{codegenerate}, text \& image question answering \cite{deepseek,imageofthought}. However, due to the limitations of the design and scale of LLMs themselves, they cannot handle very complex tasks alone \cite{llmsurvey}. AI Agent can help LLM solve more complex problems without post-training \cite{planningofllmagentsurvey}. The basic core of AI Agent is the LLMs it constructs, which can conduct conversations, complete tasks, reason based on LLMs, and can demonstrate a certain degree of autonomous behavior \cite{zhou2023agents}. The scale and design of the core LLM determine the initial capabilities and limitations of AI agents \cite{llmsurvey}.

\begin{figure}[t]
    \centering
    \includegraphics[width=0.8\linewidth]{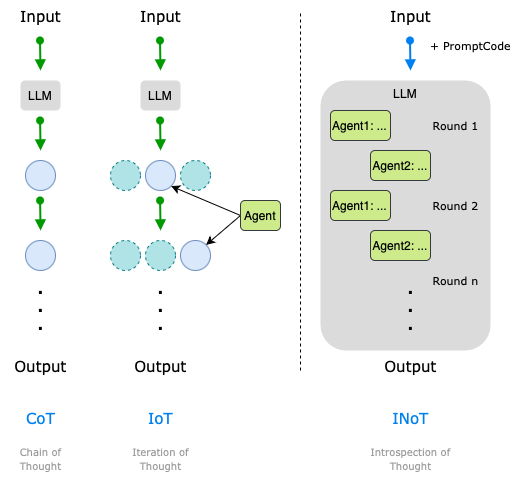}
    %\caption{Comparing to Chain of Thought (CoT) \cite{cot} and Iteration of Thought (IoT) \cite{loT} reasoning framework, our Introspection of Thought (INoT) performs an iterative inference inside the LLM itself instead of outside.}
    \label{fig:transfer}
\end{figure}

In order to improve the performance of LLM in AI Agent, prompt engineering plays an important role \cite{promptsummary,tot,loT}. Prompt Engineering seamlessly integrates a pre-trained LLM into downstream tasks without updating model parameters by inputting designed instructions, and therefore guides the model to perform the expected behavior only through the given prompts \cite{promptsummary}. Prompts can be natural language instructions that provide context to guide the model, or learning vector representations that activate relevant knowledge \cite{promptinllm,nlppromptsurvey}. AI Agents come with various built-in prompts to provide guidance, allowing LLM to perform more diverse reasoning for different tasks \cite{agentlite}. This enables AI Agents to complete more complex tasks. However, there are many ways to express the same operation in natural language. Different expressions can lead to bias in understanding in natural language, which can cause LLM not to think as expected \cite{nlppromptsurvey}. %In addition, even with guidance of prompt engineering, LLM may still output chaotic results.

To ensure that the final output of the AI Agent meets expectations, the reasoning frameworks are integrated into the AI Agent \cite{planningofllmagentsurvey}. Based on the guidance of prompt engineering, they enhance the performance of an AI Agent through various iterative methods \cite{planningofllmagentsurvey}. As shown in Figure \ref{fig:transfer}, the reasoning framework working in one single AI Agent, such as CoT, undergoes multiple chain iterative inferences in the AI Agent, reflecting and judging the results of each output of LLM until the results are considered reasonable \cite{cot}. Based on the inference of a single AI Agent, collaboration between multiple AI Agents performs better on various tasks and has become a promising method to enhance the capability of LLM, such as IoT, which allows multiple Agents to interact with each other and then iteratively generate results \cite{loT}. Through a continuous iterative dialogue, the results can be optimized. ProgCo utilizes a code-level self-correction mechanism to guide LLM for error detection and improvement through programmed feedback loops \cite{progco}. Although these methods have excellent performance in natural language tasks, they do not perform well in image interpretation and inference tasks \cite{promptsummary}. Therefore, Image-of-Thought prompting guides the MLLMs to gradually perform visual inference by automatically designing image information extraction operations, in order to improve the accuracy and interpretability of MLLM in image inference tasks \cite{imageofthought}. However, the above methods do not have great generality in multimedia inference tasks \cite{nlppromptsurvey}. Meanwhile, although the reasoning framework improves the results of LLMs, this continuous iteration consumes a lot of token cost by constantly inputting and outputting to LLM \cite{loT,tot,logicot}.

To solve the above problem, we propose a new LLM-Read code in prompt (named as PromptCode), and a novel AI Agent reasoning framework based on this prompt, named as Introspection of Thought (INoT). INoT defines and designs the LLM-Read code PromptCode in the prompt, and then put the complete reasoning logic code written by PromptCode into the prompt and make LLM reason according to the code logic. By using PromptCode reasoning logic, a virtual multi-agent debate reasoning logic code is built and executed in LLM. Therefore, self-denial and reflection occur within LLM instead of outside LLM. Our main contributions are summarized as follows.
\begin{itemize}

\item INoT uses code-integrated prompt to optimize LLM performance in zero-shot reasoning tasks by making LLM reason according to the reasoning logic code in prompt.

\item INoT puts a continuous iterative reflection process into the LLM internal through the guidance of PromptCode logic. Self-denial and reflection occur when reasoning within LLM, as shown in Figure \ref{fig:transfer}, so that the external does not need to interact with LLM multiple times, reducing the token cost of completing tasks using LLM.

\item To evaluate INoT, Seven existing reasoning frameworks are selected as baselines and six datasets are selected as benchmarks in the areas of Math, Code, and QA tasks. Our INoT outperforms existing methods by an average of 11.6\% on the benchmarks and the token cost of INoT is on average 58.3\% lower than the best-performing method in baselines. In addition, we also demonstrate the versatility of INoT in image question answering tasks through validation experiments on three Image Q \& A datasets.
\end{itemize}

\begin{figure*}[h]
    \centering
    \includegraphics[width=1\linewidth]{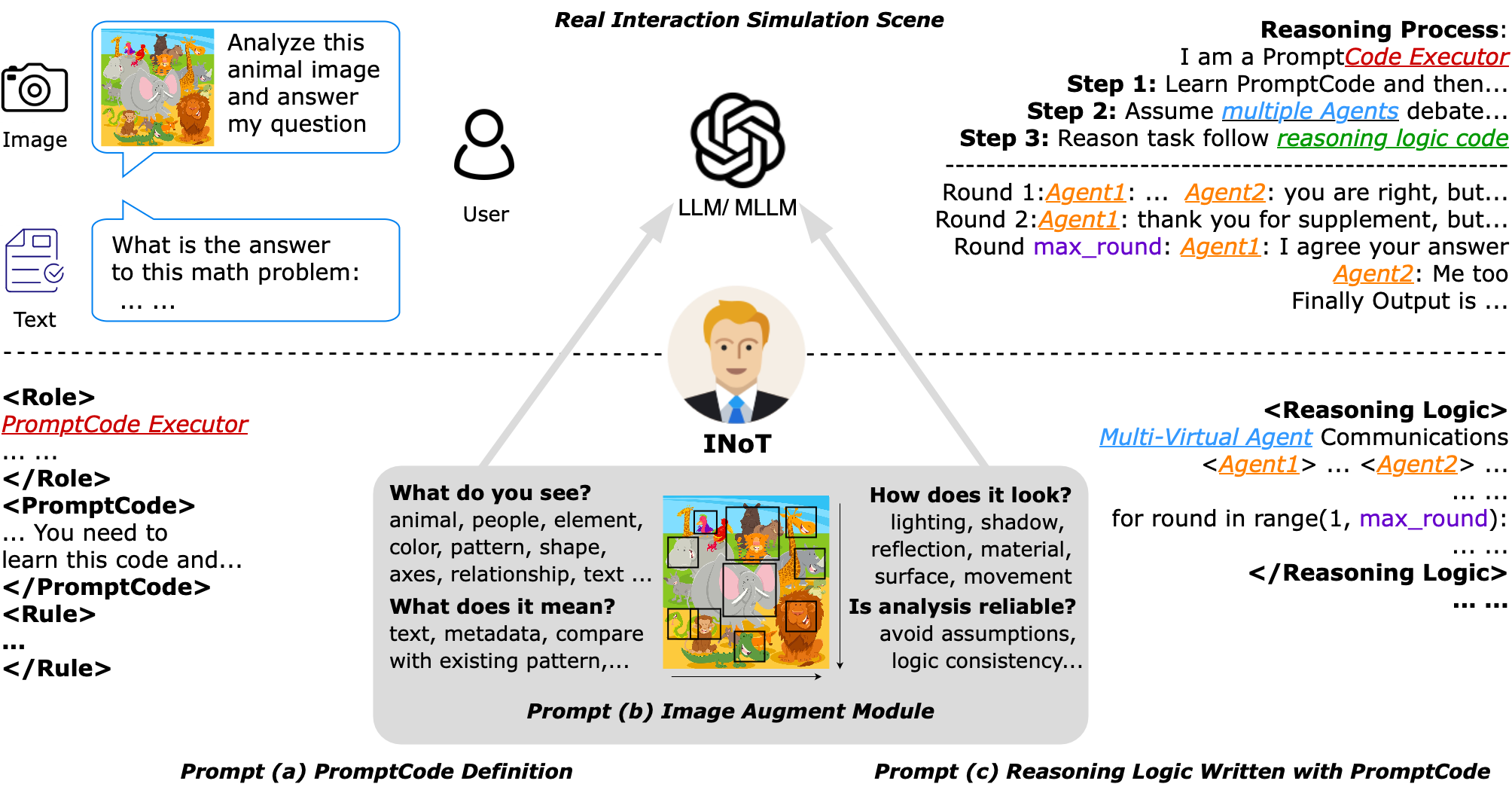}
    \caption{A multimedia inference task is combined with INoT Prompt and passed to LLM, then LLM internally processes the task according to the guidance of reasoning logic code in prompt, enabling self-denial and reflection occur within LLM instead of outside LLM. } %The upper part is a simulated LLM and human interaction scene, and the lower part is the construction of the core code-integrated prompt of INoT, where the construction of the prompt uses XML format to divide the prompt into multiple parts, each of which has a clear definition.}
    \label{fig:framework}
\end{figure*}

\section{Related Work}

\subsection{Prompt Engineering}
%Prompt Engineering has become an important technology for improving the capabilities of Large Language Models (LLMs) \cite{promptsummary}. The scale and design of the LLM determine the initial capabilities and limitations of AI Agent \cite{agentlite}. The guidance of prompts can improve the performance of LLM in AI Agent and complete specific tasks \cite{cot}. Prompt engineering guides LLM by strategically designing instructions for specific tasks, allowing LLM to generate content that meets the requirements of the task without changing the model parameters \cite{promptsummary}. By providing a mechanism to input the environment and tasks into LLM, and adjusting the model output to meet the expected results with the designed instructions, different prompt can enable LLM to perform well in various tasks \cite{cotsc}. This adaptability is different from traditional paradigms. Traditional methods often require retraining or large-scale fine-tuning of the model to achieve specific task performance \cite{parameterfunetuning,finetuninggpt}. Prompt engineering does not require a large amount of data and human input, making it more convenient than traditional methods \cite{finetuninggpt}.
Prompt engineering has become a key technique for enhancing the capability of LLM and MLLM \cite{promptsummary}. While the scale and design of LLM define its initial abilities and limitations, prompts can refine the performance of an AI Agent without modifying model parameters \cite{agentlite,cot}. By strategically designing task-specific instructions, prompt engineering enables LLMs to generate targeted outputs, adapting to various tasks \cite{promptsummary,cotsc}. Unlike traditional methods that require retraining or large-scale fine-tuning \cite{parameterfunetuning,finetuninggpt}, prompt engineering achieves task-specific improvements without extensive data or human input, offering a more efficient approach \cite{finetuninggpt}.

With the widespread application of AI Agents, research on prompt engineering continues to evolve, particularly in natural language reasoning and logic \cite{promptsummary}. Initially, tasks were directly input into LLMs using the Input-Output (IO) approach, but due to the limitations of LLM, IO could only handle simple tasks. To address more complex tasks, Chain of Thought (CoT) prompts were introduced, mimicking human reasoning to guide LLMs \cite{cot}. However, CoT was effective only for the sufficiently advanced LLMs and lacked consistency between the reasoning process and the model predictions \cite{cotsc}. To improve this problem, Self-Consistent Chain-of-Thought Distillation (SCOTT) was developed, distilling a self-consistent CoT model from a much larger teacher model \cite{cotsc}. For even more complex reasoning, LogiCoT, an instruction tuning dataset, was introduced to enhance logical reasoning and trigger general reasoning abilities in LLMs \cite{logicot}. Image-of-Thought prompt helps MLLMs better understand complex scenes by gradually extracting key visual information, optimizing the performance of MLLMs \cite{imageofthought}.
%Due to the wide application of AI Agent, research on prompt engineering is constantly developing in various aspects, especially in natural language reasoning and logic \cite{promptsummary}. Initially, humans directly input tasks to LLMs (Input-Output, IO), but due to the limitations of LLMs themselves, IO can only handle some simple tasks. In order to complete slightly more complex tasks, people introduced Chain of Thought (CoT) prompts, which imitate human thinking and gradually reason to prompt LLM \cite{cot}. However, CoT is only suitable for some sufficiently LLMs, and there is a lack of consistency between the generated reasoning process and the model's prediction \cite{cotsc}. Therefore, Self-Consistent Chain-Of Thought Distillation (SCOTT) was proposed, which learns a small and self-consistent CoT model from a teacher model several orders of magnitude larger \cite{cotsc}. In order to handle more complex tasks, some people have proposed LogiCoT, an instruction tuning dataset for Logical Chain-of-Thought Reasoning, which is used to teach models logical reasoning and trigger their general reasoning abilities \cite{logicot}. 

Furthermore, Iteration of Thought (IoT) framework was proposed to enhance the responsiveness of LLM by generating "thought-provoking" prompts for input queries and current iterative responses of LLMs \cite{loT}. Program of Thoughts (PoT) prompt  significantly improves the accuracy and efficiency of numerical reasoning tasks by separating complex computational tasks and delegating them to external program interpreters, solving the problem of computational errors in traditional methods \cite{programofthoughts}. The Program-driven Self-Correction (ProgCo) method achieves self-correction of the model by automatically generating and executing verification programs, significantly improving the inference performance in complex tasks and reducing the misleading feedback of errors \cite{progco}. PoT and ProgCo combine programming ideas to optimize LLM, but they both generate real code and execute it in an external real environment, failing to make LLM truly programmatic inference. The above prompts are designed to optimize LLM or MLLM by constructing prompts for a certain task in natural language or images. There is no prompt that can universally optimize the performance of both LLM and MLLM. In addition, due to the limitations of understanding of natural language, LLM may not be able to strictly follow the guidance in natural language prompts for reasoning \cite{promptsummary}. Therefore, in terms of guiding LLM reasoning, directly constructing the code for reasoning logic and putting it into the prompt, which will make the LLM strictly follow the code logic for reasoning and ensure the results more rigorous and accurate.

\subsection{AI Agent}
The advancement of LLMs has accelerated the progress of AI Agents \cite{llmsurvey,planningofllmagentsurvey}. AI Agents extend beyond simple text generation by leveraging LLMs as their core computing engine, enabling dialogue, task execution, reasoning, and a degree of autonomy \cite{dynamicllmpoweredagentnetwork,planningofllmagentsurvey}. Carefully designed prompts encode identity, instructions, authorization, and context to shape agent behavior \cite{llmsurvey}. While LLMs form the foundation of AI Agents, inference strategies and agent architectures are crucial \cite{agentlite}. Research has evolved from static, single-agent thought-chain prompts to dynamic, multi-agent dialogue systems \cite{loT,cot}.

Some people focus on optimizing a prompt to enable LLMs to complete specific tasks, while others utilize different proxy architectures to optimize the performance of AI Agents. For example, an AI proxy library AgentLite was proposed, which simplified the innovation process of inference strategies, proxy architectures, and applications by providing a lightweight and user-friendly platform \cite{agentlite}. Instead of using static methods, a framework called Dynamic LLM-Powered Agent Network (DyLAN) was built, which collaborated on different tasks and domains with a dynamic communication structure for LLM-driven proxy collaboration, operating a two-stage paradigm: (1) team optimization, and (2) task resolution \cite{dynamicllmpoweredagentnetwork}.

All the above methods interact with only one single LLM. In addition, multiple agents have been developed to communicate with each other for reasoning. For example, Iteration of Thought (IoT) framework, which enhanced the responsiveness of LLM by generating thought-provoking prompts based on input queries and the current iterative response of LLM \cite{loT}. However, the token cost of current methods designed for the Agent architecture was extremely high. The final result depended on iterative optimization outside the LLM or dialogue between multiple LLMs \cite{loT,agentlite,tot}. These reflection processes required continuous inputs and outputs from LLMs, which generated a mass of inference costs \cite{planningofllmagentsurvey}. If there is a method that can incorporate external iterative and dialogue processes into the LLM, it will save a lot of token costs.

\section{Method}

We propose the Introspection of Thought (INoT) framework, which defines a LLM-read code, PromptCode, in the prompt, as shown in Figure \ref{fig:framework}. We use PromptCode to write reasoning logic code, and put the complete definition of PromptCode and reasoning logic code into the prompt, guiding LLM to strictly follow the code logic in the prompt for inference. This transfers the self-denial and reflection process of conventional AI Agents from outside LLM to inside LLM, thereby improving LLM inference performance while reducing token cost.

After investigating each prompt structure, we choose XML as the overall designing framework for prompts in INoT, using the XML structure to construct prompt. XML uses a hierarchical tree structure to organize data, dividing it into different levels and nodes through nested tags. This structure can clearly express the logical relationship between each part of prompt, providing LLM with an intuitive and easy-to-parse prompt framework, improving parsing efficiency of LLM, and reducing misunderstandings caused by non-standard or ambiguous prompt formats. At the same time, XML tags are self-descriptive, and tag names are usually defined based on the content or meaning of the section, which makes the section itself has strong semantic expression ability. When LLM processes prompt, it can quickly understand the meaning of each prompt fragment without relying on additional contextual information, thus significantly improving the accuracy and efficiency of its semantic understanding.

There are three modules in the INoT prompt, the PromptCode Definition Module, the Augment Module, and the Reasoning Module.

\begin{lstlisting}[caption={PromptCode Definition (XML as the Language)}, label={role}, escapeinside={(*@}{@*)}]
(*@\textbf{<Role>}@*)
RoleName: PromptCode Executor
RoleDesc: You need to learn PromptCode,  
a structured reasoning code defined below. 
Follow the rules and execute reasoning logic 
code strictly as written in <Reasoning Logic>. 
(*@\textbf{</Role>}@*)
(*@\textbf{<PromptCode>}@*)
PromptCode is a structured reasoning code  
that explicitly defines logical steps  
to solve a given task. It is a hybrid of 
Python programming and natural language.
(*@\textbf{</PromptCode>}@*)
(*@\textbf{<Rule>}@*)
Purpose of each module designed below:    
<Image Augment>:Follow this when analyse 
image, if no image, ignore this. 
<Reasoning Logic>: The most important part, 
must reason task follow this line by line!
(*@\textbf{</Rule>}@*)
\end{lstlisting}

\subsection{PromptCode Definition Module}

PromptCode is a new program language defined in INoT prompt, specifically designed for LLM to read and understand. In order to better enable LLM to learn and understand, PromptCode is composed of a mixture of Python programming language and natural language. Unlike pseudocode, pseudocode is mainly designed for human understanding, while PromptCode is designed specifically to cater to semantic parsing ability and generation logic of LLM, in order to facilitate LLM reading and understanding.

%Python has a concise and clear syntax, a clear code structure, and can express complex task logic and steps in a highly logical way. Python code itself contains rich semantic information, such as variable names, function names, and comments, which provides LLM with additional contextual clues and helps them better understand the intention of the code. Natural language can describe the background, goals, and requirements of the task in an intuitive and flexible way, enabling LLM to quickly grasp the overall intention of the task. As Python programming language is close to natural language, it is not abrupt to integrate it with natural language. When Python provides LLM with clear and rigorous logical instructions, natural language can further provide LLM with rich semantic clues, helping them better understand the semantic background and details of the task. At the same time, natural language can replace some complex Python code logic, especially in scenarios involving complex task descriptions and dynamic logic adjustments. In some cases, only one sentence of natural language can represent dozens or even hundreds of lines of Python code, which can effectively make the reasoning logic code in prompts more concise and thus be efficiently understood by LLM.
Python features a concise syntax and clear code structure, enabling the efficient expression of complex task logic. Its code inherently contains rich semantic information, such as variable names, function names, and comments, which provides LLMs with additional contextual cues to better infer code intent. Meanwhile, natural language offers an intuitive and flexible means of conveying task background, goals, and requirements, allowing LLMs to grasp the overall task intention more effectively. Due to the linguistic proximity between Python and natural language, their integration is seamless and mutually reinforcing. While Python provides precise logical instructions, natural language supplements them with rich semantic context, enhancing the model’s understanding of task details. In scenarios involving complex task descriptions or dynamic logic adjustments, natural language can often replace lengthy code blocks, expressing the same logic in a single sentence. This significantly improves the conciseness and interpretability of reasoning prompts for LLMs.

%In addition, integrating natural language into Python can more effectively stimulate the potential of LLM. In prompt, natural language can describe some functions that are difficult to directly implement in Python in reality, which often have high abstraction, dynamism, or involve complex logical reasoning. Through the description of natural language, LLM can break through the limitations of traditional programming languages when understanding tasks, thereby generating more creative and adaptive solutions. Python code usually requires clear logic and structure to implement specific functions, but may face difficulties in dealing with highly abstract concepts. For example, describing nonfunctional requirements such as "scalability" and "robustness" of a system, or describing the style of images, it is difficult to directly write code in Python. However, by describing these abstract concepts in natural language, LLM can better understand the core requirements of the task and generate output that meet these requirements.

Integrating natural language into Python can further enhance the capabilities of LLMs. In prompt, natural language can describe some functions that are difficult to directly implement in Python in reality, which often have high abstraction, dynamism, or involve complex logical reasoning. Through the description of natural language, LLM can break through the limitations of traditional programming languages when understanding tasks, thereby generating more creative and adaptive solutions. Python code usually requires clear logic and structure to implement specific functions, but may face difficulties in dealing with highly abstract concepts. For instance, non-functional requirements like “scalability” and “robustness,” or stylistic attributes of images, are challenging to encode directly in code. However, by expressing these concepts in natural language, LLMs can better grasp the core requirements and generate outputs aligned with them.

In PromptCode Definition, Rule enables LLM to understand the entire prompt clearly and efficiently. Rule provides a brief introduction to all modules in the prompt in advance, as shown in Listing \ref{role}, which can construct a clear reasoning framework and execution specifications in prompt, guiding LLM to follow specific logical structures for reasoning and decision-making in different task scenarios. It ensures that LLM can think systematically when dealing with complex tasks, avoid arbitrary speculation, and improve the accuracy, consistency, and controllability of reasoning.

\begin{lstlisting}[caption={Image Augment}, label={imageaugment}, escapeinside={(*@}{@*)}]

(*@\textbf{<Image Augment>}@*)  
Follow this module to analyze images.   
(*@\textbf{<Basic Visual Understanding>}@*)
(What do you see?)  
- Identify people, animals, key elements.  
- Recognize colors, shapes, patterns, axes, 
legends and spatial relationships.  
- Detect text within images (if present) 
(*@\textbf{</Basic Visual Understanding>}@*)
(*@\textbf{<Advanced Visual Analysis>}@*)
(How does it look?)  
- Observe lighting, shadows, and reflections. 
- Analyze textures, materials, and surfaces. 
- Consider movement cues (if applicable) 
(*@\textbf{</Advanced Visual Analysis>}@*)
(*@\textbf{<Context Awareness>}@*)
(What does it mean?)  
- Use any accompanying text, metadata, or 
task instructions to refine understanding.  
- Compare the image with known patterns or 
knowledge to improve accuracy.  
- Recognize symbolic or cultural references 
that may impact interpretation.  
(*@\textbf{</Context Awareness>}@*)
(*@\textbf{<Inference and Verification>}@*)
(Is the analysis reliable?)  
- Cross-check image details with available 
textual descriptions.  
- Avoid assumptions if something is unclear,
acknowledge uncertainty instead of guessing.  
- Maintain logical consistency: ensure your 
visual interpretation aligns with input task.  
(*@\textbf{</Inference and Verification>}@*)
(*@\textbf{</Image Augment>}@*)  


\end{lstlisting}

\subsection{Image Augment Module}

Image Augment module, as part of the INoT prompt, aims to guide MLLM to follow a systematic visual analysis framework when processing tasks that involve images to ensure precision, consistency, and logic of understanding, as shown in Listing \ref{imageaugment}. When the input task contains images, the Image Augment Module will play a role. This module uses clear instructions to enable MLLM to perform basic visual perception when parsing images, identify key elements, colors, shapes, patterns, axes, legends, spatial relationships, and textual information, thereby establishing a preliminary understanding of the image content. Subsequently, the prompt further requires LLM to conduct a more in-depth visual analysis, focusing on environmental factors such as lighting, shadows, and reflections, in order to infer time, material, and physical characteristics, and combine dynamic features to infer possible time-series information. At the same time, this module clearly instructs LLM to combine contextual information, including task descriptions, text annotations, and existing knowledge, to improve recognition accuracy, understand possible cultural or symbolic meanings, and avoid one-sided reasoning. In addition, this prompt module requires LLM to conduct strict logical verification during the reasoning process to ensure that the analysis results conform to the context and task objectives, avoid arbitrary guessing, and express appropriate uncertainty when the information is uncertain. Through this series of precise guidelines, the Image Augment module, as part of the INoT prompt, not only standardizes the way LLM processes images, but also enhances its depth of visual understanding, enabling it to provide more reliable and reasonable reasoning results in various task scenarios.

\subsection{Reasoning Module} % Logic Execution

Reasoning Logic module in INoT prompt is designed to structure a multi-agent debate mechanism within the reasoning process of LLM, ensuring that responses are reflected and refined iteratively through critical evaluation and self-denial. The reasoning framework assumes the presence of two independent agents, Agent$\_$A and Agent$\_$B, who engage in a structured debate to iteratively refine their responses. The debate proceeds through several structured phases: initial reasoning, argument presentation, critique, rebuttal, and adjustment, ultimately leading to supervised validation and output of the final response.

\begin{lstlisting}[caption={Reasoning Code}, label={reasoning}, escapeinside={(*@}{@*)}]
(*@\textbf{<ReasoningLogic>}@*)
(*@\textbf{if}@*) image (*@\textbf{in}@*) task: import (*@\textbf{<Image Augment>}@*)
#Assume agents with independent reasoning.
Agent_A/B = (*@\textbf{DebateAgent(task)}@*)
result_A, thought_A = Agent_A.reason()
result_B, thought_B = Agent_B.reason()
# Set debate parameters
MaxRounds=10,Counter=0,agreement=False
(*@\textbf{While}@*) not agreement or Counter < MaxRounds:
Counter += 1
At each step, you must think like Agent_A 
and then like Agent_B, responding as if 
two intelligent debaters are reflecting.
#(*@\textbf{Step 1:Agents present their arguments}@*)
argument_A = Agent_A.reason()
argument_B = Agent_B.reason()
#(*@\textbf{Step 2: Critique each other's reasoning}@*)
(*@\textbf{(Debate Phase)}@*)
critique_A = Agent_A.critique(argument_B)  
# A critiques B's reasoning
critique_B = Agent_B.critique(argument_A)  
# B critiques A's reasoning
#(*@\textbf{Step 3: Agents respond to critiques}@*)
(*@\textbf{(Rebuttal Phase)}@*)
rebuttal_A = Agent_A.rebut(critique_B)  
# A rebuts B's critique
rebuttal_B = Agent_B.rebut(critique_A)  
# B rebuts A's critique
#(*@\textbf{Step 4: Adjust based on rebuttals}@*)
(*@\textbf{(Adjustment Phase)}@*)
result_A,thought_A=Agent_A.adjust(rebuttal_B)  
# A refines based on B's rebuttal
result_B,thought_B=Agent_B.adjust(rebuttal_A)  
# B refines based on A's rebuttal
#(*@\textbf{Step 5: Agreement Check}@*)
agreement = (result_A == result_B)
final_result = result_A if agreement 
Output final_result without explanation.
(*@\textbf{</ReasoningLogic>}@*)

\end{lstlisting}

The whole reasoning logic code is shown in Listing \ref{reasoning}. The process begins by incorporating an Image Augment module to enhance interpretability before reasoning. Then, LLM begins to initialize Agent$\_$A and Agent$\_$B, both tasked with reasoning independently on the given problem. Each agent produces an initial result along with the corresponding thought process. The debate mechanism is regulated through a maximum number of rounds (MaxRounds = 10) to prevent infinite loops, while a Counter ensures structured iteration. The core of the process lies in the debate, where in each round, both agents present their arguments, engage in mutual critique, respond through rebuttals, and adjust their reasoning accordingly. This iterative debate-rebuttal-adjustment cycle enables both agents to refine their thoughts based on opposing viewpoints, thereby reducing reasoning errors and inconsistencies. If both agents reach an agreement, their refined results converge and the result is finalized. Otherwise, the debate continues until either an agreement is reached or the iteration limit is met. The final response is then presented without additional explanations, reinforcing the structured reasoning approach and leading to more reliable output.

%\noindent The reasoning logic code is shown in {\color{red}Alog }. The reflection mechanism first utilizes Bayesian reasoning to enable multiple agents to share information and adjust their respective reasoning paths during the reasoning process. The update of Bayesian reasoning can reflect the information exchange and collaboration between these agents. Meanwhile, the use of graph neural networks (GNN) provides an effective information dissemination mechanism for multi-agent systems, allowing each agent to integrate information from neighboring agents in each iteration, thereby improving decision accuracy.

%\begin{equation}\label{3}
%\begin{aligned}
%h_i^{(k+1)} = \text{Aggregate}\left( \{ h_j^{(k)} | j \in \mathcal{N}(i) \} \right) + W^{(k)} h_i^{(k)}

%\end{aligned}
%\end{equation}

%\noindent$H_{i} (k)$ is the hidden representation of the i-th node (Agent) at the k-th layer. \mathcal {N} (i) is the set of neighbor nodes of the i-th node. $W ^ { (k) }$ is the weight matrix of the kth layer, used to transform the features of node i. $\text {Aggregate} (\ cdot)$ is an information aggregation function that usually fuses information from neighboring nodes through weighted averaging or other aggregation methods. Each agent exchanges information with other agents during the inference process and updates its own state through graph neural networks. In each iteration, the agent updates its inference based on the state of other agents.

\begin{table*}[t]
    \centering

  \begin{tabular}{c|cccccc|c}
    \toprule  
    \multirow{2}*{\textbf{Methods}} & \textbf{HumanEval}&\textbf{MBPP} &\textbf{MATH}& \textbf{GSM8K}&\textbf{HotpotQA}&\textbf{SQuAD}
    &\multirow{2}*{\textbf{Avg.}} \\

       \cline{2-7}
    &pass$@$1 &pass$@$1 &Solve Rate(\%) &Solve Rate(\%) &F1 Score  &F1 Score \\
    
       \toprule
  \multirow{1}*{IO} &88.2&72.2&48.2&90.0&65.2&84.2&74.7\\

    \multirow{1}*{CoT} &89.3&72.2&49.2&89.2&63.7&85.3&74.8\\
    
    \multirow{1}*{SCCOT} &89.7&73.3&48.9&90.8&65.1&86.7&75.8\\
    
    \multirow{1}*{LogiCoT} &89.7&73.7&48.7&90.2&64.3&85.2&75.3\\
    
    \multirow{1}*{ToT} &90.6&74.2&50.2&90.4&66.5&86.2&76.4\\
    
    \multirow{1}*{GIoT} &90.2&72.2&47.5&90.4&63.3&85.1&74.8\\
    
    \multirow{1}*{AIoT} &90.4&72.8&47.2&91.2&64.3&86.3&75.4\\
    \toprule
    \multirow{1}*{Ours} &\textbf{95.9}&\textbf{84.3}&\textbf{53.4}&\textbf{93.3}&\textbf{72.2}&\textbf{88.8}&\textbf{81.3}\\
    \toprule

\end{tabular}

 \caption{Comparative analysis  of performance of different methods and INoT on divided test set of Code, Math, QA tasks and average performance of each methods. The methods in this experiment uniformly use DeepSeek-V2.5. In the table, the best-performing results on the same dataset test set are highlighted in bold.}
 \label{tab:baseline}
\end{table*}

\begin{figure*}[t]
    \centering
    \includegraphics[scale = 1.0, height=0.4\linewidth, width=1\linewidth]{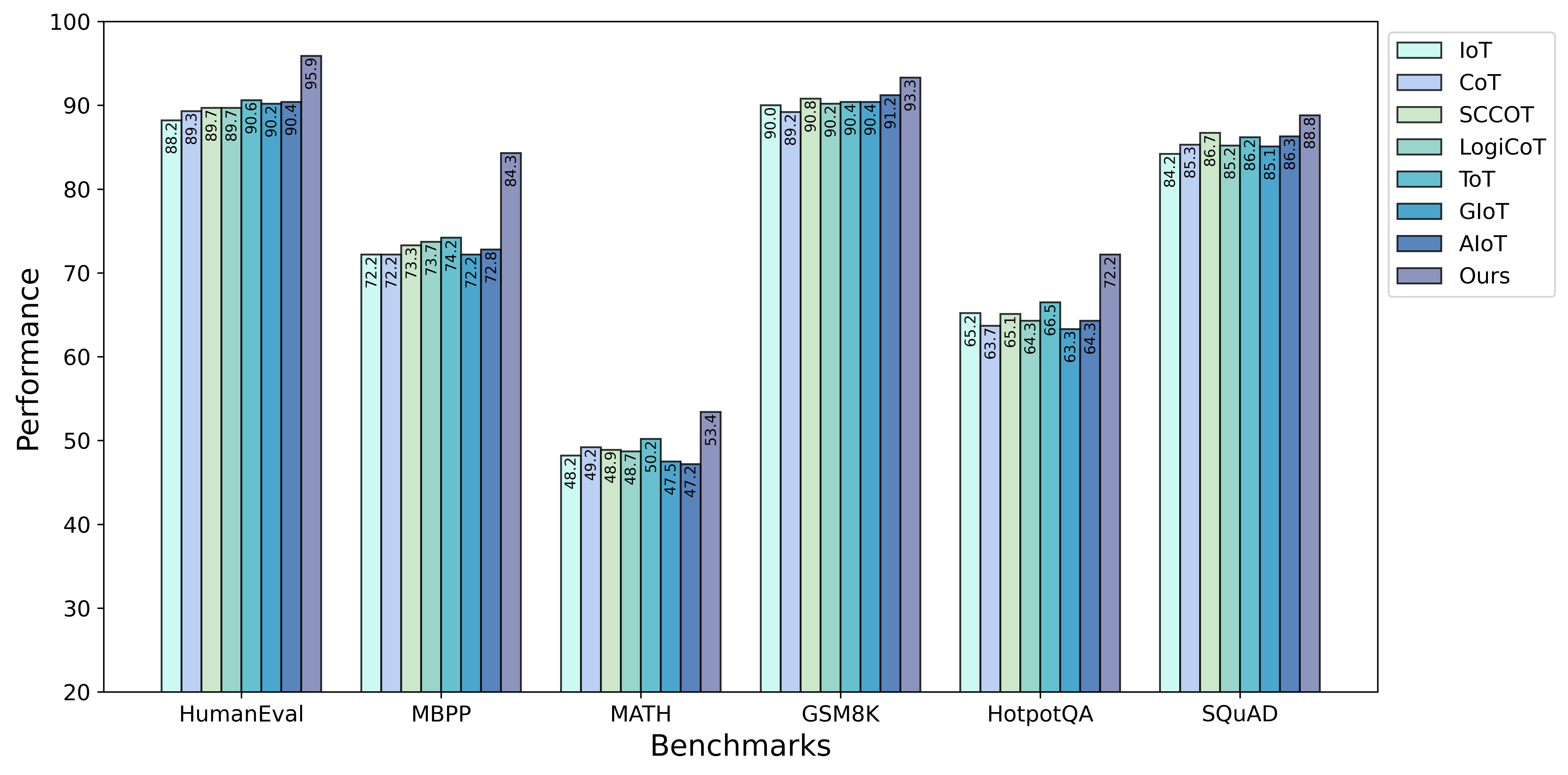}
    \caption{\textbf{Performance comparison with baselines. }Experiments are conducted on six datasets representing Math, Code, and QA tasks. INoT outperforms baselines on each dataset.}
    \label{fig:benchmark}
\end{figure*}

%The generative and supervisory agents in LLM obtain information from Bayesian inference and graph neural networks to achieve iterative optimization and reflection. 
%In addition, by combining the attention and memory mechanisms, a dynamic evaluation mechanism is designed during the evaluation and reward process. This mechanism allows LLM to dynamically adjust its reasoning direction and strategy based on historical status, task progress, and feedback from other agents.

%\begin{equation}\label{3}
%\begin{aligned}
%V_i(t) = \sum_{j=1}^{n} \alpha_{i,j}(t) \cdot M_j(t-1) + \beta_i \cdot f_i(x_t)

%\end{aligned}
%\end{equation}

%\noindent$V_{i} (t)$ is the evaluation value of the supervision Agent for the $i-$th Agent at time $t$, indicating the overall evaluation of the current state. $\alpha_{i, j}(t)$ is the attention weight of the $i-$th Agent to the memory state $M_j (t-1) $ of the $j-$th agent. $M_j (t-1)$ is the memory state of the $j-$th agent at time $t-1$, reflecting the reasoning history of other agents. $\beta_i $is the weight of the $i-$th Agent's current state, used to control the influence of its own reasoning process. $f_i (x_t)$ is the $i-$th Agent's response to the input $x_t$.

\begin{table*}[h]
    \centering

  \begin{tabular}{c|cccccc}
    \toprule  
    \multirow{2}*{\textbf{Models}}  
    & \textbf{HumanEval}&\textbf{MBPP} &\textbf{MATH}& \textbf{GSM8K}&\textbf{HotpotQA}&\textbf{SQuAD}
    
\\
\cline{2-7}
    &pass$@$1 &pass$@$1 &Solve Rate(\%) &Solve Rate(\%) &F1 Score  &F1 Score \\
    
       \toprule
  \multirow{1}*{DeepSeek-V2.5} &95.9&84.3&53.4&93.3&72.2&88.8\\

    \multirow{1}*{DeepSeek-V2} &93.4&82.4&52.8&90.2&71.6&87.4\\\
    
    \multirow{1}*{Claude-3.5-sonnet} &96.2&84.5&53.2&93.8&72.1&86.7\\
    
    \multirow{1}*{Gemma 2} &93.2&82.5&52.2&90.3&72.0&84.8\\
    
    \multirow{1}*{Qwen2.5-coder} &92.7&81.7&51.7&91.2&70.3&84.2\\
    
    \multirow{1}*{Llama3.2} &93.3&79.4&51.2&90.4&71.5&84.3\\
    
    \toprule
  
\end{tabular}

 \caption{Comparative analysis of performance of INoT when utilizing different LLMs as the core LLM.}
\label{tab:differentmodel}
\end{table*}

%This formula combines the dynamic evaluation of each Agent in the inference process and focuses on the historical inference state $M_j (t-1)$ of other agents and its own immediate response $f_i (x_t)$ through the attention mechanism, so that LLM assigns a comprehensive evaluation value $V_i (t)$ to each agent. The evaluation value reflect the Agent's current contribution and performance in the entire inference process.

%\begin{equation}\label{3}
%\begin{aligned}
%\mathcal{M}_{t+1} = \sum_{i=1}^{n} \alpha_i \cdot \mathcal{R}_i
%\end{aligned}
%\end{equation}

%\noindent$M_{t + 1}$ is the comprehensive information shared by each agent at time $t + 1$. $\alpha_i$ is the weight of each agent, representing the degree of influence of the Agent's result on general reasoning. $\alpha_i$ satisfies $\sum_ {i = 1} ^ {n}\alpha_i = 1$. $\mathcal {R} _i $ is the reasoning result obtained by Agent $A_{i}$ at time $t$. In the inference logic code, the inference results of all agents $\mathcal {R} _i$ are weighted and fused according to the weight $\alpha_i $ to generate comprehensive inference information at time $t + 1$. Generally, each generated agent has the same weight and the supervision Agent has a higher weight, which is twice that of the generated Agent.

\section{Experiments}

\subsection{Setup}
\textbf{Datasets. }We select three types of tasks to evaluate the performance of INoT, including QA, Code and Math domains. Six public benchmarks are used for these tasks for our experiments. According to established practices for addressing these three tasks, the data is divided into validation and test sets using a $1:4$ ratio \cite{aflow}. Specifically, in QA domain, HotpotQA \cite{hotpotqa} and SQuAD 1.1 \cite{rajpurkar-etal-2016-squad} are selected, where we randomly select 1,000 samples each. In Code domain, we use the Python dataset for HumanEval \cite{humaneval} and full dataset for MBPP \cite{mbpp}. In terms of Math, we use the entire dataset for GSM8K \cite{GSM8K} to refer to existing work \cite{cot}, and for MATH \cite{math}, 600 random problems with difficulty levels of 4 and 5 are selected from three types of problems (Combinatorics \& Probability, Pre-algebra, Pre-calculus). Furthermore, in multimedia inference versatility verification, three image QA datasets are selected as benchmarks to refer to existing work \cite{hu2025socraticquestioninglearnselfguide}, which are ScienceQA-IMG \cite{scienceqa}, LLaVA-Bench (COCO) \cite{llava}, and LLaVA-Bench (In-the-Wild) \cite{llava}. 
\\

\noindent \textbf{Implementation Details. }INoT utilizes different LLMs for tasks. LLMs including DeepSeek-V2.5 \cite{deepseek}, DeepSeek-Chat in DeepSeek-V2 \cite{deepseek}, Claude-3.5-sonnet \cite{claude}, Gemma 2 \cite{gemma}, Qwen2.5-coder \cite{qwen} and Llama3.2 \cite{llama} are employed as core LLMs. In addition, LLaVA \cite{llava}, LLaMA-Adapter \cite{llamaadapter}, and Multimodal Chain-of-Thought (MM-CoT) \cite{mmcot} are selected in multimedia inference versatility verification. All models are accessed via APIs. We set the temperature at 1 for DeepSeek-V2.5 and 0 for the other models. We set iteration rounds to 10 in INoT prompt. The environment configuration for experiment is: Pytorch 2.1.2 + cuda 12.1 as frame, Intel ® Core ™ i9-14900KF as CPU and NVIDIA GeForce RTX 4090 as GPU.\\
\\

\noindent \textbf{Baselines. }In the experiments, we compare the results generated by INoT on the three types of tasks with other reasoning prompt engineering methods \cite{promptsummary}. These methods include IO (direct LLM invocation), Chain-of-Thought (CoT) \cite{cot}, Self-Consistency Chain-of-Thought (SCCOT) \cite{cotsc}, Logical Cain-of-Thought (LogiCoT) \cite{logicot}, Tree-of-Thought (ToT) \cite{tot}, guided iteration of thought (GIoT) and autonomous iteration of thought (AIoT) \cite{loT}. In these methods, LLMs are uniformly implemented using DeepSeek-V2.5 \cite{deepseek}. Additionally, in order to verify the universality and effectiveness of INoT in image inference tasks, we compared the results of four methods on benchmarks. These methods are IO, CoT, INoT without Image Augment Module (w/o IAM), and complete INoT on benchmarks. \\
\\

\begin{figure*}[t]
    \centering
    \includegraphics[width=0.9\linewidth]{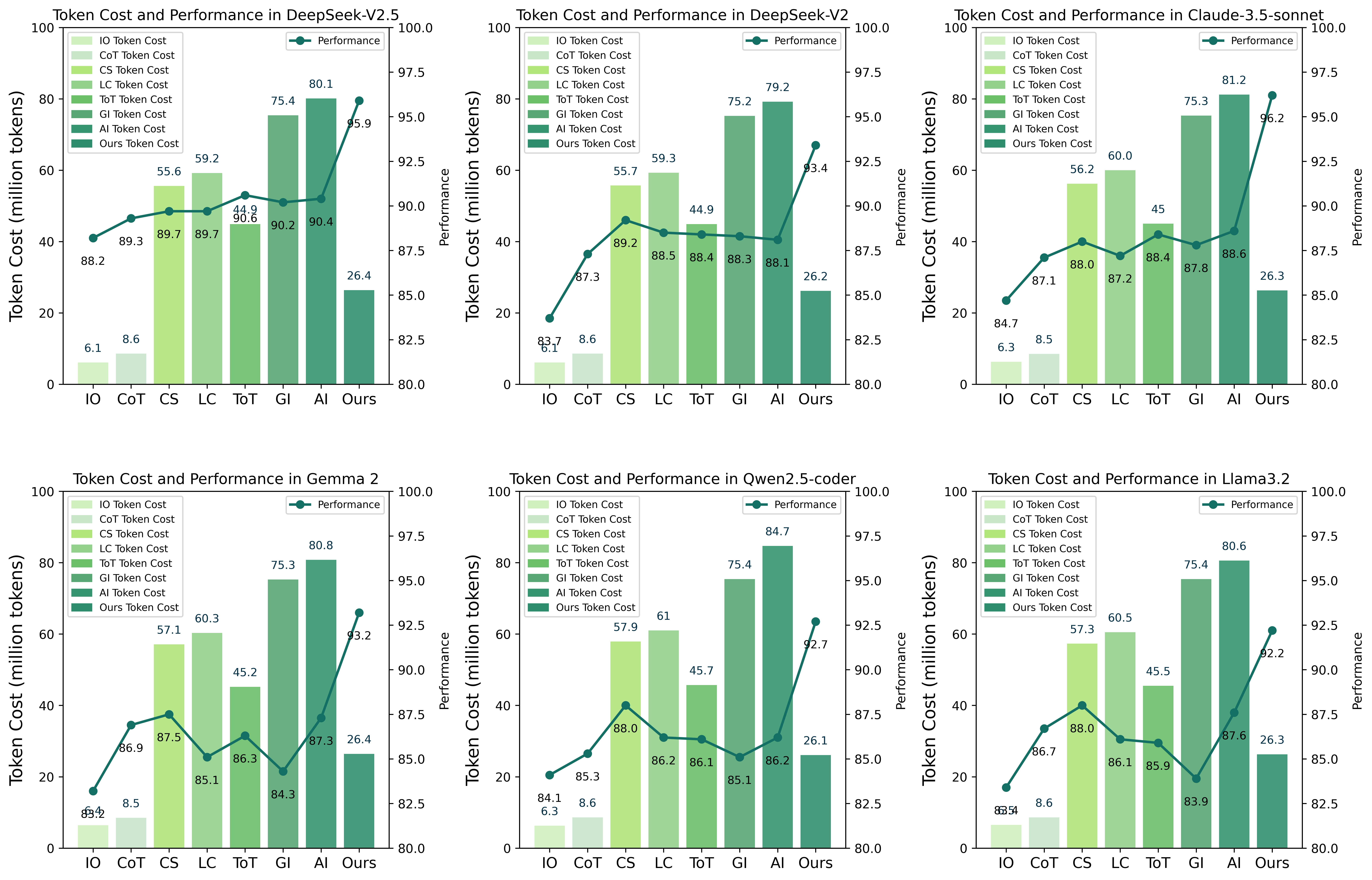}
    \caption{\textbf{Analysis of token cost. }Comparing the token cost and performance of baselines and INoT on the HumanEval dataset test set utilizing different LLMs. In this chart, CS represents SCCOT, LC represents LogiCoT, GI represents GIoT and AI represents AIoT. The line chart shows the performance, and the bar chart shows the token cost.}
    \label{fig:cost_analyse}
\end{figure*}

 \noindent\textbf{Evaluation Metrics. }For the QA tasks, we report the F1 Score as the metric. For the Code tasks, we report the pass$@$1 metric following established practices \cite{archon} to evaluate the accuracy of the generated code. For the Math tasks, we report the Solve Rate (\%). All the final result is the average of five experiments. Moreover, we construct a pareto chart to analyze the cost of executing tasks on each test set by tracking token usage. This approach visually demonstrates the performance-cost trade-offs across different methods.

\subsection{Results and Analysis }

\noindent \textbf{Comparison to Baselines. }Experimental results on six datasets in QA, Code, Math domains, as shown in Table \ref{tab:baseline}, reflect the effectiveness and applicability of INoT compared to the baseline methods. INoT achieves an average performance of 81.3, surpassing all baseline methods by an average margin of 7.95\%. This consistent improvement highlights INoT’s ability to enhance model reasoning and problem-solving across diverse domains. Specifically, INoT outperforms the best-performing baseline, ToT, by 6.41 percentage points in overall performance. INoT exhibits substantial gains in code generation tasks, achieving a pass@1 score of 95.9 on HumanEval and 84.3 on MBPP, significantly higher than the best baseline scores of 90.6 and 74.2, respectively. Similarly, in mathematical problem-solving, INoT attains a solve rate of 53.4 on MATH and 93.3 on GSM8K, surpassing the best baseline performance of 50.2 and 91.2. For QA benchmarks, INoT achieves an F1 score of 75.2 on HotpotQA and 87.8 on SQuAD, representing notable improvements over the best baseline performances of 66.5 and 86.7, respectively. These results underscore INoT’s robustness and adaptability across different problem types, affirming its advantage in text task reasoning and structured problem-solving scenarios.\\

%INoT achieved an average performance of 82.7\% on the benchmarks, which outperform all the baselines methods by an average of 11.16\%. Meanwhile, the favorable performance across the three tasks demonstrates that INoT is adaptable to different types of tasks.\\

\noindent \textbf{Comparative Study Among Different LLMs. }As shown in Table \ref{tab:differentmodel}, the performance of INoT across different LLMs demonstrates its model-agnostic nature. We conduct experiments with INoT using different LLMs on benchmarks. On the same datasets, the performance variation of INoT across different models does not exceed 5\%. The results indicate that INoT shows stable performance and is not significantly affected by the variations in LLMs.
\\

\begin{figure}[t]
    \centering
    \includegraphics[width=1\linewidth]{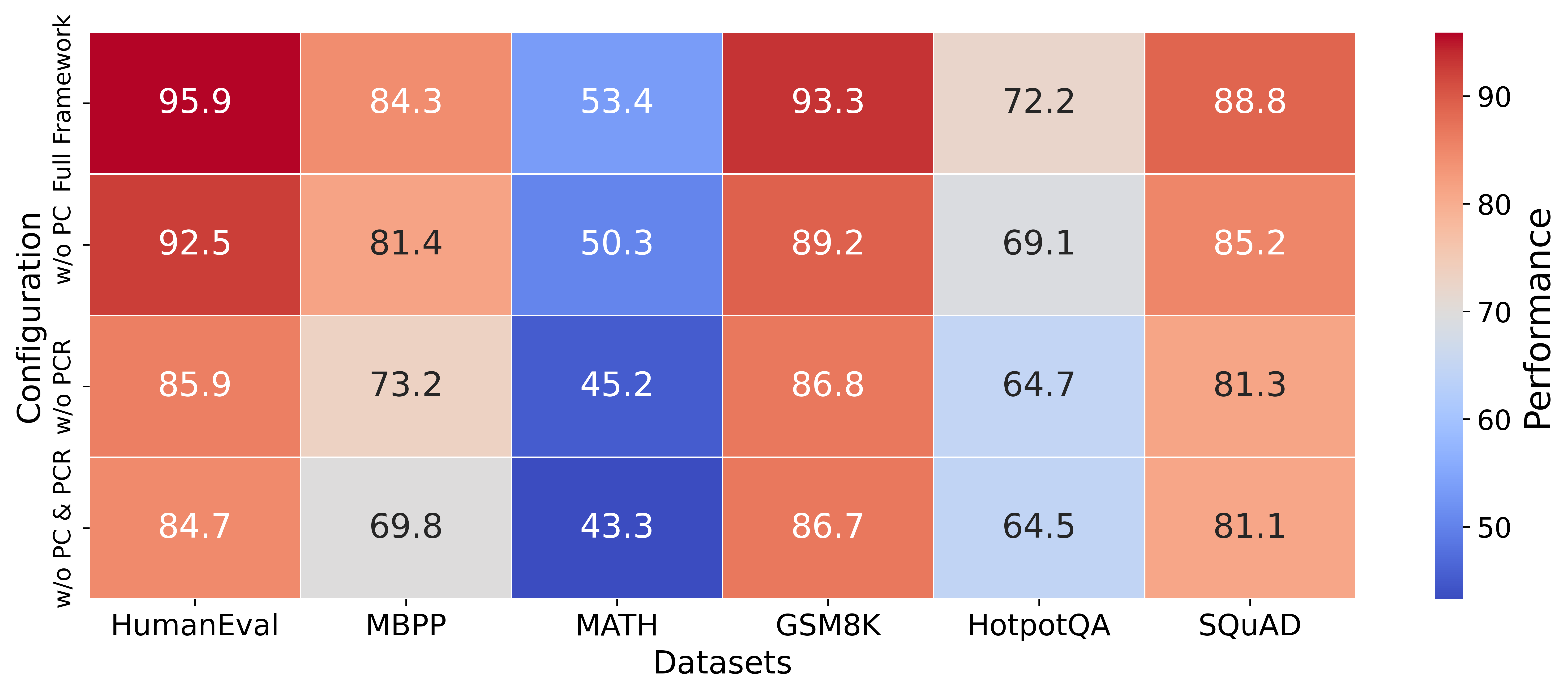}
    %\caption{\textbf{Ablation experiment results.} INoT conducts experiments on benchmarks in four situations using DeepSeek-V2.5. In this figure, PC indicates PromptCode definition in INoT prompt, and PCR is PromptCode reasoning logic code in INoT prompt.}
    \label{fig:ablation}
\end{figure}

\noindent \textbf{Cost Analysis. }As the Figure \ref{fig:cost_analyse} shown, we compare the performance and token cost between the baselines and INoT. The range of token cost ranges from a minimum of 6.1 to a maximum of 84.7, with significant differences. The token cost of INoT in all systems remains at a low level, usually around 26, far below the average level of other methods. Token cost of INoT is 58.3\% lower on average than the best performing method in baselines. From the comprehensive comparison of token cost and performance score, the "Ours" method shows significant advantages. It not only has the lowest token cost, but also has the highest performance score, showing the best balance between resource efficiency and task effectiveness.\\
\\

\begin{table}[t]
    \centering
 
  \begin{tabular}{c|c|ccc}
    \toprule  
   \multirow{1}{1cm}{\textbf{Methods}} & \multirow{1}*{\textbf{Setting}} 
    & \textbf{SQA-IMG} &\textbf{LLaVA\textsuperscript{qa}} &\textbf{LLaVA\textsuperscript{w}}
    
       \\
      % \cline{3-5}
    %&&Accuracy &Accuracy &Accuracy  \\
    
       \toprule
  \multirow{4}{1cm}{LLaVA}& IO&88.0&85.1&67.3\\

                 &CoT&88.2&80.2&68.2\\
    
                &ours(w/o IAM)&88.9&81.2&69.8\\
    
                &ours &\textbf{90.2}&\textbf{83.4}&\textbf{72.4}\\

    \toprule
     \multirow{4}{1cm}{LLaMA-Adaptor}& IO&80.3&76.7&62.4\\

                 &CoT&81.2&77.2&62.8\\
    
                &ours(w/o IAM)&81.8&78.6&63.9\\
    
                &ours &\textbf{83.9}&\textbf{80.2}&\textbf{70.1}\\
    \toprule

     \multirow{4}{1cm}{MM-CoT}& IO&82.9&77.6&63.9\\

                 &CoT&83.2&78.1&64.2\\
    
                &ours(w/o IAM)&83.8&79.2&64.7\\
    
                &ours &\textbf{85.7}&\textbf{80.4}&\textbf{68.7}\\
    \toprule

\end{tabular}
  
 \caption{Accuracy (\%) on three image QA benchmarks. SQA-IMG: ScienceQA-IMG (zero-shot); LLaVA\textsuperscript{qa}: LLaVA-Bench (COCO); LLaVA\textsuperscript{w}: LLaVA-Bench (In-the-Wild). INoT outperforms baselines on each benchmarks. Additionally, Image Augment Module in INoT prompt is verified to be useful. }
 \label{tab:imageverify}
\end{table}

\noindent \textbf{Multimedia Versatility Analysis. }Experimental results on three image QA benchmarks, as shown in Table \ref{tab:imageverify}, demonstrate that INoT has excellent performance compared to various base lines. Under different settings and benchmarks, the proposed method consistently outperforms the baselines, proving the effectiveness and versatility of INoT in image inference tasks. Compared with the base line methods (IO and CoT), the proposed method has an average accuracy improvement of about 2.7\% on SQA-IMG; an average improvement of about 4.5\% on LLaVA-Bench (COCO); and an average improvement of about 6.0\% on LLaVA-Bench (In-the-Wild). These improvements indicate that the proposed method can effectively enhance the performance of the model in image QA tasks, especially on the more challenging LLaVA-Bench (In-the-Wild) dataset, where the improvement is more significant. The LLaMA-Adaptor and MM-CoT models have also achieved significant improvements in their respective benchmark tests, further demonstrating the generalization and adaptability of the method. In addition, the results also verify the usefulness of the Image Augment Module in INoT prompt. INoT without Image Augment Module shows lower accuracy on all benchmarks of each model than the full INoT. For example, in LLaVA, the accuracy of INoT without Image Augment Module on SQA-IMG is 88.9\%, the accuracy on LLaVA-Bench (COCO) is 81.2\%, and the accuracy on LLaVA-Bench (In-the-Wild) is 69.8\%, which is 1.3\%, 2.2\%, and 2.6\% lower than INoT, respectively.
\\

\noindent \textbf{Ablation Study. }In order to verify whether the custom PromptCode in the code-integrated prompt really works, we conduct ablation experiments on benchmarks. As shown in Figure \ref{fig:ablation}, these are the experimental results after removing the custom code of PromptCode. On the same dataset, the performance of INoT has a certain degree of decline. Additionally, it can be seen that the impact of the execution of the PromptCode logic is greater than that of the design of PromptCode and PromptComplier, indicating that running PromptCode to specify the reasoning logic plays a significant role in the optimal performance of LLMs.

\section{Conclusion and Future Work } % and Future Work
We propose a novel AI Agent Reasoning Framework with Introspection of Thought (INoT) by designing a new LLM-Read code in prompt. It constructs a virtual multi-agent debate reasoning framework within LLM. Therefore, self-denial and reflection occur within LLM instead of outside LLM. Benchmark experiments are conducted on Math, Code, and QA tasks. The performance of INoT exceeds baselines, demonstrating its effectiveness in completing natural language tasks. In addition, putting the iterative reflection and optimization of the conventional agent reasoning framework into LLM effectively reduces the cost of completing natural language tasks. Furthermore, we demonstrate the versatility of INoT in image interpretation and inference through verification experiments on three image QA benchmarks. For future work, we would like to continue to optimize the code-integrated prompts, enhance the internal reasoning logic of LLM. Also, we will apply the INoT reasoning framework to more diverse tasks. %{\color{red}Future work}

%%%%%%%%%%%%%%%%%%%%%%%%%%%%%%%%%%%%%%%%%%%%%%%%%%%%%%%%%%%%%%%%%%%%%%%%

%%% Use this command to include your bibliography file.

\bibliography{main}

\begin{thebibliography}{38}
\providecommand{\natexlab}[1]{#1}
\providecommand{\url}[1]{\texttt{#1}}
\expandafter\ifx\csname urlstyle\endcsname\relax
  \providecommand{\doi}[1]{doi: #1}\else
  \providecommand{\doi}{doi: \begingroup \urlstyle{rm}\Url}\fi

\bibitem[Anthropic(2024)]{claude}
Anthropic.
\newblock \emph{Claude-3.2-sonnet}.
\newblock 2024.
\newblock URL \url{https://www.anthropic.com/news/claude-3-5-sonnet}.

\bibitem[Austin et~al.(2021)Austin, Odena, Nye, Bosma, Michalewski, Dohan, Jiang, Cai, Terry, Le, and Sutton]{mbpp}
J.~Austin, A.~Odena, M.~Nye, M.~Bosma, H.~Michalewski, D.~Dohan, E.~Jiang, C.~Cai, M.~Terry, Q.~Le, and C.~Sutton.
\newblock Program synthesis with large language models, 2021.
\newblock URL \url{https://arxiv.org/abs/2108.07732}.

\bibitem[Chen et~al.(2021)Chen, Tworek, Jun, Yuan, de~Oliveira~Pinto, Kaplan, Edwards, Burda, Joseph, Brockman, Ray, Puri, Krueger, Petrov, Khlaaf, Sastry, Mishkin, Chan, Gray, Ryder, Pavlov, Power, Kaiser, Bavarian, Winter, Tillet, Such, Cummings, Plappert, Chantzis, Barnes, Herbert-Voss, Guss, Nichol, Paino, Tezak, Tang, Babuschkin, Balaji, Jain, Saunders, Hesse, Carr, Leike, Achiam, Misra, Morikawa, Radford, Knight, Brundage, Murati, Mayer, Welinder, McGrew, Amodei, McCandlish, Sutskever, and Zaremba]{humaneval}
M.~Chen, J.~Tworek, H.~Jun, Q.~Yuan, H.~P. de~Oliveira~Pinto, J.~Kaplan, H.~Edwards, Y.~Burda, N.~Joseph, G.~Brockman, A.~Ray, R.~Puri, G.~Krueger, M.~Petrov, H.~Khlaaf, G.~Sastry, P.~Mishkin, B.~Chan, S.~Gray, N.~Ryder, M.~Pavlov, A.~Power, L.~Kaiser, M.~Bavarian, C.~Winter, P.~Tillet, F.~P. Such, D.~Cummings, M.~Plappert, F.~Chantzis, E.~Barnes, A.~Herbert-Voss, W.~H. Guss, A.~Nichol, A.~Paino, N.~Tezak, J.~Tang, I.~Babuschkin, S.~Balaji, S.~Jain, W.~Saunders, C.~Hesse, A.~N. Carr, J.~Leike, J.~Achiam, V.~Misra, E.~Morikawa, A.~Radford, M.~Knight, M.~Brundage, M.~Murati, K.~Mayer, P.~Welinder, B.~McGrew, D.~Amodei, S.~McCandlish, I.~Sutskever, and W.~Zaremba.
\newblock Evaluating large language models trained on code, 2021.
\newblock URL \url{https://arxiv.org/abs/2107.03374}.

\bibitem[Chen et~al.(2023)Chen, Ma, Wang, and Cohen]{programofthoughts}
W.~Chen, X.~Ma, X.~Wang, and W.~W. Cohen.
\newblock Program of thoughts prompting: Disentangling computation from reasoning for numerical reasoning tasks, 2023.
\newblock URL \url{https://arxiv.org/abs/2211.12588}.

\bibitem[Cobbe et~al.(2021)Cobbe, Kosaraju, Bavarian, Chen, Jun, Kaiser, Plappert, Tworek, Hilton, Nakano, Hesse, and Schulman]{GSM8K}
K.~Cobbe, V.~Kosaraju, M.~Bavarian, M.~Chen, H.~Jun, L.~Kaiser, M.~Plappert, J.~Tworek, J.~Hilton, R.~Nakano, C.~Hesse, and J.~Schulman.
\newblock Training verifiers to solve math word problems, 2021.
\newblock URL \url{https://arxiv.org/abs/2110.14168}.

\bibitem[DeepSeek(2024)]{deepseek}
DeepSeek.
\newblock \emph{DeepSeek-V2.5}.
\newblock 2024.
\newblock URL \url{https://huggingface.co/deepseek-ai/DeepSeek-V2.5}.

\bibitem[Ding et~al.(2023)Ding, Qin, Yang, Wei, Yang, Su, Hu, Chen, Chan, Chen, et~al.]{parameterfunetuning}
N.~Ding, Y.~Qin, G.~Yang, F.~Wei, Z.~Yang, Y.~Su, S.~Hu, Y.~Chen, C.-M. Chan, W.~Chen, et~al.
\newblock Parameter-efficient fine-tuning of large-scale pre-trained language models.
\newblock \emph{Nature Machine Intelligence}, 5\penalty0 (3):\penalty0 220--235, 2023.

\bibitem[Dong et~al.(2024)Dong, Jiang, Jin, and Li]{codegenerate}
Y.~Dong, X.~Jiang, Z.~Jin, and G.~Li.
\newblock Self-collaboration code generation via chatgpt.
\newblock \emph{ACM Transactions on Software Engineering and Methodology}, 33\penalty0 (7):\penalty0 1--38, 2024.

\bibitem[Gemma~Team et~al.(2024)Gemma~Team, Hardin, Dadashi, Bhupatiraju, Sifre, Rivière, Kale, Love, Tafti, Hussenot, and et~al.]{gemma}
T.~M. Gemma~Team, C.~Hardin, R.~Dadashi, S.~Bhupatiraju, L.~Sifre, M.~Rivière, M.~S. Kale, J.~Love, P.~Tafti, L.~Hussenot, and et~al.
\newblock Gemma.
\newblock 2024.
\newblock \doi{10.34740/KAGGLE/M/3301}.
\newblock URL \url{https://www.kaggle.com/m/3301}.

\bibitem[Hendrycks et~al.(2021)Hendrycks, Burns, Kadavath, Arora, Basart, Tang, Song, and Steinhardt]{math}
D.~Hendrycks, C.~Burns, S.~Kadavath, A.~Arora, S.~Basart, E.~Tang, D.~Song, and J.~Steinhardt.
\newblock Measuring mathematical problem solving with the math dataset, 2021.
\newblock URL \url{https://arxiv.org/abs/2103.03874}.

\bibitem[Hu et~al.(2025)Hu, Liu, Chen, Zhou, Xiao, Yang, and Zhang]{hu2025socraticquestioninglearnselfguide}
W.~Hu, H.~Liu, L.~Chen, F.~Zhou, C.~Xiao, Q.~Yang, and C.~Zhang.
\newblock Socratic questioning: Learn to self-guide multimodal reasoning in the wild, 2025.
\newblock URL \url{https://arxiv.org/abs/2501.02964}.

\bibitem[Huang et~al.(2024)Huang, Liu, Chen, Wang, Wang, Lian, Wang, Tang, and Chen]{planningofllmagentsurvey}
X.~Huang, W.~Liu, X.~Chen, X.~Wang, H.~Wang, D.~Lian, Y.~Wang, R.~Tang, and E.~Chen.
\newblock Understanding the planning of llm agents: A survey.
\newblock \emph{arXiv preprint arXiv:2402.02716}, 2024.

\bibitem[Latif and Zhai(2024)]{finetuninggpt}
E.~Latif and X.~Zhai.
\newblock Fine-tuning chatgpt for automatic scoring.
\newblock \emph{Computers and Education: Artificial Intelligence}, 6:\penalty0 100210, 2024.

\bibitem[Liu et~al.(2023{\natexlab{a}})Liu, Li, Wu, and Lee]{llava}
H.~Liu, C.~Li, Q.~Wu, and Y.~J. Lee.
\newblock Visual instruction tuning, 2023{\natexlab{a}}.
\newblock URL \url{https://arxiv.org/abs/2304.08485}.

\bibitem[Liu et~al.(2023{\natexlab{b}})Liu, Teng, Cui, Zhang, Zhou, and Zhang]{logicot}
H.~Liu, Z.~Teng, L.~Cui, C.~Zhang, Q.~Zhou, and Y.~Zhang.
\newblock Logicot: Logical chain-of-thought instruction-tuning, 2023{\natexlab{b}}.
\newblock URL \url{https://arxiv.org/abs/2305.12147}.

\bibitem[Liu et~al.(2023{\natexlab{c}})Liu, Yuan, Fu, Jiang, Hayashi, and Neubig]{nlppromptsurvey}
P.~Liu, W.~Yuan, J.~Fu, Z.~Jiang, H.~Hayashi, and G.~Neubig.
\newblock Pre-train, prompt, and predict: A systematic survey of prompting methods in natural language processing.
\newblock \emph{ACM Computing Surveys}, 55\penalty0 (9):\penalty0 1--35, 2023{\natexlab{c}}.

\bibitem[Liu et~al.(2024{\natexlab{a}})Liu, Yao, Zhang, Yang, Liu, Tan, Choubey, Lan, Wu, Wang, et~al.]{agentlite}
Z.~Liu, W.~Yao, J.~Zhang, L.~Yang, Z.~Liu, J.~Tan, P.~K. Choubey, T.~Lan, J.~Wu, H.~Wang, et~al.
\newblock Agentlite: A lightweight library for building and advancing task-oriented llm agent system.
\newblock \emph{arXiv preprint arXiv:2402.15538}, 2024{\natexlab{a}}.

\bibitem[Liu et~al.(2024{\natexlab{b}})Liu, Zhang, Li, Liu, and Yang]{dynamicllmpoweredagentnetwork}
Z.~Liu, Y.~Zhang, P.~Li, Y.~Liu, and D.~Yang.
\newblock A dynamic llm-powered agent network for task-oriented agent collaboration, 2024{\natexlab{b}}.
\newblock URL \url{https://arxiv.org/abs/2310.02170}.

\bibitem[Llama(2024)]{llama}
Llama.
\newblock \emph{Llama3.2}.
\newblock 2024.
\newblock URL \url{https://ollama.com/library/llama3.2}.

\bibitem[Long(2023)]{tot}
J.~Long.
\newblock Large language model guided tree-of-thought, 2023.
\newblock URL \url{https://arxiv.org/abs/2305.08291}.

\bibitem[Lu et~al.(2022)Lu, Mishra, Xia, Qiu, Chang, Zhu, Tafjord, Clark, and Kalyan]{scienceqa}
P.~Lu, S.~Mishra, T.~Xia, L.~Qiu, K.-W. Chang, S.-C. Zhu, O.~Tafjord, P.~Clark, and A.~Kalyan.
\newblock Learn to explain: Multimodal reasoning via thought chains for science question answering.
\newblock \emph{Advances in Neural Information Processing Systems}, 35:\penalty0 2507--2521, 2022.

\bibitem[Marvin et~al.(2023)Marvin, Hellen, Jjingo, and Nakatumba-Nabende]{promptinllm}
G.~Marvin, N.~Hellen, D.~Jjingo, and J.~Nakatumba-Nabende.
\newblock Prompt engineering in large language models.
\newblock In \emph{International conference on data intelligence and cognitive informatics}, pages 387--402. Springer, 2023.

\bibitem[Qwen(2024)]{qwen}
Qwen.
\newblock \emph{Qwen2.5-coder}.
\newblock 2024.
\newblock URL \url{https://ollama.com/library/qwen2.5-coder}.

\bibitem[Radha et~al.(2024)Radha, Jelyani, Ghukasyan, and Goktas]{loT}
S.~K. Radha, Y.~N. Jelyani, A.~Ghukasyan, and O.~Goktas.
\newblock Iteration of thought: Leveraging inner dialogue for autonomous large language model reasoning, 2024.
\newblock URL \url{https://arxiv.org/abs/2409.12618}.

\bibitem[Rajpurkar et~al.(2016)Rajpurkar, Zhang, Lopyrev, and Liang]{rajpurkar-etal-2016-squad}
P.~Rajpurkar, J.~Zhang, K.~Lopyrev, and P.~Liang.
\newblock {SQ}u{AD}: 100,000+ questions for machine comprehension of text.
\newblock In J.~Su, K.~Duh, and X.~Carreras, editors, \emph{Proceedings of the 2016 Conference on Empirical Methods in Natural Language Processing}, pages 2383--2392, Austin, Texas, Nov. 2016. Association for Computational Linguistics.
\newblock \doi{10.18653/v1/D16-1264}.
\newblock URL \url{https://aclanthology.org/D16-1264}.

\bibitem[Roumeliotis and Tselikas(2023)]{chatgptreview}
K.~I. Roumeliotis and N.~D. Tselikas.
\newblock Chatgpt and open-ai models: A preliminary review.
\newblock \emph{Future Internet}, 15\penalty0 (6):\penalty0 192, 2023.

\bibitem[Saad-Falcon et~al.(2024)Saad-Falcon, Lafuente, Natarajan, Maru, Todorov, Guha, Buchanan, Chen, Guha, R{\'e}, et~al.]{archon}
J.~Saad-Falcon, A.~G. Lafuente, S.~Natarajan, N.~Maru, H.~Todorov, E.~Guha, E.~K. Buchanan, M.~Chen, N.~Guha, C.~R{\'e}, et~al.
\newblock Archon: An architecture search framework for inference-time techniques.
\newblock \emph{arXiv preprint arXiv:2409.15254}, 2024.

\bibitem[Sahoo et~al.(2024)Sahoo, Singh, Saha, Jain, Mondal, and Chadha]{promptsummary}
P.~Sahoo, A.~K. Singh, S.~Saha, V.~Jain, S.~Mondal, and A.~Chadha.
\newblock A systematic survey of prompt engineering in large language models: Techniques and applications, 2024.
\newblock URL \url{https://arxiv.org/abs/2402.07927}.

\bibitem[Song et~al.(2025)Song, Wu, Wang, Liu, Su, and Zheng]{progco}
X.~Song, Y.~Wu, W.~Wang, J.~Liu, W.~Su, and B.~Zheng.
\newblock Progco: Program helps self-correction of large language models, 2025.
\newblock URL \url{https://arxiv.org/abs/2501.01264}.

\bibitem[Wang et~al.(2023)Wang, Wang, Li, Gao, Yin, and Ren]{cotsc}
P.~Wang, Z.~Wang, Z.~Li, Y.~Gao, B.~Yin, and X.~Ren.
\newblock Scott: Self-consistent chain-of-thought distillation, 2023.
\newblock URL \url{https://arxiv.org/abs/2305.01879}.

\bibitem[Wei et~al.(2022)Wei, Wang, Schuurmans, Bosma, ichter, Xia, Chi, Le, and Zhou]{cot}
J.~Wei, X.~Wang, D.~Schuurmans, M.~Bosma, b.~ichter, F.~Xia, E.~Chi, Q.~V. Le, and D.~Zhou.
\newblock Chain-of-thought prompting elicits reasoning in large language models.
\newblock In S.~Koyejo, S.~Mohamed, A.~Agarwal, D.~Belgrave, K.~Cho, and A.~Oh, editors, \emph{Advances in Neural Information Processing Systems}, volume~35, pages 24824--24837. Curran Associates, Inc., 2022.
\newblock URL \url{https://proceedings.neurips.cc/paper_files/paper/2022/file/9d5609613524ecf4f15af0f7b31abca4-Paper-Conference.pdf}.

\bibitem[Yang et~al.(2018)Yang, Qi, Zhang, Bengio, Cohen, Salakhutdinov, and Manning]{hotpotqa}
Z.~Yang, P.~Qi, S.~Zhang, Y.~Bengio, W.~W. Cohen, R.~Salakhutdinov, and C.~D. Manning.
\newblock Hotpotqa: A dataset for diverse, explainable multi-hop question answering, 2018.
\newblock URL \url{https://arxiv.org/abs/1809.09600}.

\bibitem[Yao et~al.(2024)Yao, Duan, Xu, Cai, Sun, and Zhang]{llmsurvey}
Y.~Yao, J.~Duan, K.~Xu, Y.~Cai, Z.~Sun, and Y.~Zhang.
\newblock A survey on large language model (llm) security and privacy: The good, the bad, and the ugly.
\newblock \emph{High-Confidence Computing}, page 100211, 2024.

\bibitem[Zhang et~al.(2024{\natexlab{a}})Zhang, Xiang, Yu, Teng, Chen, Chen, Zhuge, Cheng, Hong, Wang, Zheng, Liu, Luo, and Wu]{aflow}
J.~Zhang, J.~Xiang, Z.~Yu, F.~Teng, X.~Chen, J.~Chen, M.~Zhuge, X.~Cheng, S.~Hong, J.~Wang, B.~Zheng, B.~Liu, Y.~Luo, and C.~Wu.
\newblock Aflow: Automating agentic workflow generation, 2024{\natexlab{a}}.
\newblock URL \url{https://arxiv.org/abs/2410.10762}.

\bibitem[Zhang et~al.(2023)Zhang, Han, Liu, Gao, Zhou, Hu, Yan, Lu, Li, and Qiao]{llamaadapter}
R.~Zhang, J.~Han, C.~Liu, P.~Gao, A.~Zhou, X.~Hu, S.~Yan, P.~Lu, H.~Li, and Y.~Qiao.
\newblock Llama-adapter: Efficient fine-tuning of language models with zero-init attention.
\newblock \emph{arXiv preprint arXiv:2303.16199}, 2023.

\bibitem[Zhang et~al.(2024{\natexlab{b}})Zhang, Zhang, Li, Zhao, Karypis, and Smola]{mmcot}
Z.~Zhang, A.~Zhang, M.~Li, H.~Zhao, G.~Karypis, and A.~Smola.
\newblock Multimodal chain-of-thought reasoning in language models, 2024{\natexlab{b}}.
\newblock URL \url{https://arxiv.org/abs/2302.00923}.

\bibitem[Zhou et~al.(2024)Zhou, Zhou, Hu, Lu, Gao, and Zhang]{imageofthought}
Q.~Zhou, R.~Zhou, Z.~Hu, P.~Lu, S.~Gao, and Y.~Zhang.
\newblock Image-of-thought prompting for visual reasoning refinement in multimodal large language models.
\newblock \emph{arXiv preprint arXiv:2405.13872}, 2024.

\bibitem[Zhou et~al.(2023)Zhou, Jiang, Li, Wu, Wang, Qiu, Zhang, Chen, Wu, Wang, et~al.]{zhou2023agents}
W.~Zhou, Y.~E. Jiang, L.~Li, J.~Wu, T.~Wang, S.~Qiu, J.~Zhang, J.~Chen, R.~Wu, S.~Wang, et~al.
\newblock Agents: An open-source framework for autonomous language agents.
\newblock \emph{arXiv preprint arXiv:2309.07870}, 2023.

\end{thebibliography}

\end{document}